\documentclass[conference]{IEEEtran}
\IEEEoverridecommandlockouts
\usepackage{cite}
\usepackage{amsmath,amssymb,amsfonts}
\usepackage{algorithmic}
\usepackage{graphicx}
\usepackage{textcomp}
\usepackage{xcolor}
\usepackage{cuted}
\usepackage{subcaption}
\def\BibTeX{{\rm B\kern-.05em{\sc i\kern-.025em b}\kern-.08em
    T\kern-.1667em\lower.7ex\hbox{E}\kern-.125emX}}
\begin{document}

\title{A Hybrid Controller Design for Human-Assistive Piloting of an Underactuated Blimp 
}

\author{\IEEEauthorblockN{Wugang Meng}
\IEEEauthorblockA{\textit{Department of Electronic and Computer Engineering} \\
\textit{Hong Kong University of Science and Technology}\\
Hong Kong, China \\
0000-0003-4289-2409}
\and
\IEEEauthorblockN{Tianfu Wu}
\IEEEauthorblockA{\textit{Department of Electronic and Computer Engineering} \\
\textit{Hong Kong University of Science and Technology}\\
Hong Kong, China \\
0000-0001-7683-777X}
\and
\IEEEauthorblockN{Qiuyang Tao}
\IEEEauthorblockA{\textit{ China} \\
qiuyang.tao@outlook.com \\
0000-0002-3413-1311}
\and
\IEEEauthorblockN{Fumin Zhang}
\IEEEauthorblockA{\textit{Department of Electronic and Computer Engineering} \\
\textit{Department of Mechanical and Aerospace Engineering} \\
\textit{Hong Kong University of Science and Technology}\\
Hong Kong, China \\
0000-0003-0053-4224}
}


\maketitle

\begin{abstract}{

This paper introduces a novel solution to the manual control challenge for indoor blimps. The problem's complexity arises from the conflicting demands of executing human commands while maintaining stability through automatic control for underactuated robots. To tackle this challenge, we introduced an assisted piloting hybrid controller with a preemptive mechanism, that seamlessly switches between executing human commands and activating automatic stabilization control. Our algorithm ensures that the automatic stabilization controller operates within the time delay between human observation and perception, providing assistance to the driver in a way that remains imperceptible.
}
\end{abstract}


\section{Introduction}
Unmanned aerial vehicles (UAVs) have become increasingly popular in various fields including military, agriculture, and transportation. However, the high cost\cite{mccarley2004human} of UAV pilot training\cite{weeks2000unmanned} and the difficulty of executing tasks indoors \cite{khosiawan2019task,khosiawan2016system} have limited their widespread use. Blimps, on the other hand, have been proven to be safe and suitable for indoor flight\cite{palossi2019nano,ferdous2019developing,bang}. Indoor blimps have made significant advancements in various applications like surveillance and inspection\cite{sampedro2019fully}.
And previous research\cite{cho2017autopilot, palossi2019nano,ferdous2019developing, bang} on indoor blimps has centered primarily on autonomous flight, little attention has been given to human-computer interaction from the operator's perspective. This paper aims to propose a hybrid controller design to assist human pilots in driving a blimp indoors.

\begin{figure}[h]
    \centerline{\includegraphics[width=0.45\textwidth]{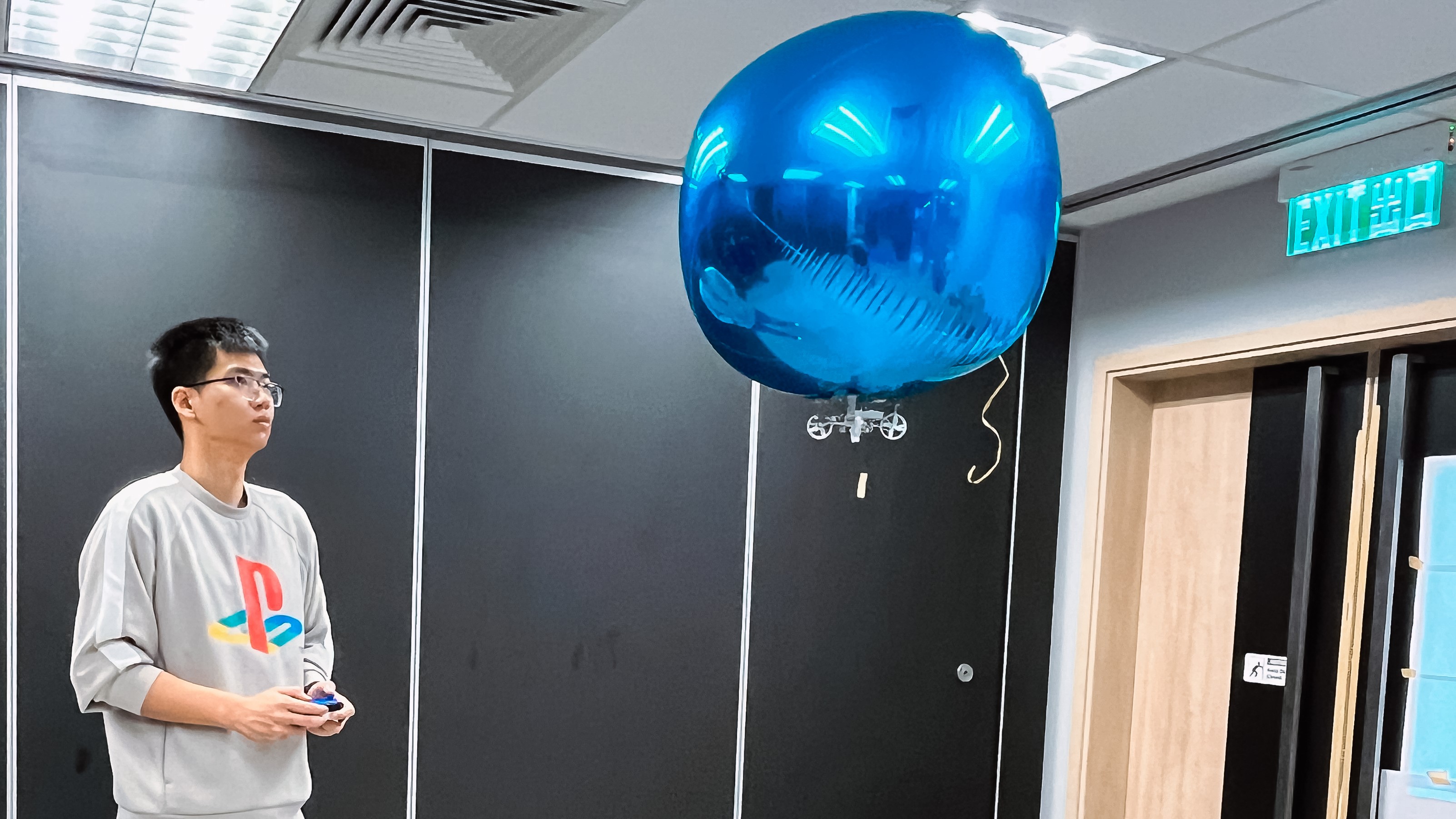}}
    \caption{Photograph of a human operator operating an indoor miniature blimp. }
    \label{fig:human}
\end{figure}
Wegner proposed three principles that play a crucial role in shaping the human operator experience in the context of human-computer interaction \cite{WEGNER200365}. These principles are: “(1) priority: conscious intention to perform an act must immediately precede the action, which should precede the outcome; (2) consistency: “the sensory outcome must fit the predicted outcome”; and, (3) exclusivity: “one’s thoughts must be the only apparent cause of the outcome”. Jeunet et al. demonstrated that violations of these principles can significantly impact human perceptions of event causation and subsequently affect the overall operational experience \cite{8260944}. For underactuated robots like blimps, the execution of human commands and the maintenance of stable flight through automated control pose conflicting challenges. The high-frequency automated control commands has the potential to override the human's commands, creating a false perception of loss of control and leading to incorrect human inputs. 

To address this conflict, we have developed a preemptive mechanism that allows the blimp to seamlessly switch between executing human commands and activating automatic stabilization. This mechanism takes into account the time delay between human observation and event detection. Our algorithm ensures that the automatic stabilizing controller is only active within this time lag and ends before the operator is able to detect it. Our proposed hybrid controller design, coupled with the preemptive mechanism, provides stabilizing flight assistance to blimp pilots without the need for external sensors nor pre-flight aerodynamic parameter measurements. This innovative approach offers a novel solution for indoor blimp control.

The significance of our work lies in the fact that it provides a cost-effective feasible solution to the problem of more comfortable human piloting of indoor blimps, which can have a wide range of applications in various fields such as surveillance, search and rescue, and entertainment. Our hybrid controller enables the blimp to accurately execute commands from the human pilot while maintaining stable flight conditions, without violating Wegner's three principles. By adhering to these principles, we aim to create a seamless and user-friendly interaction between the blimp and the human operator. The proposed hybrid controller design is a  contribution to the human-robot interaction, and can pave the way for further research in this area.


\section{Background on Blimp Design}\label{design}
\subsection{Actuation Design}\label{AA}

Conventional indoor miniature blimps should prioritize multi-directional maneuverability, which requires obtaining the exact position and sailing direction of the blimp. However, the indoor electromagnetic environment is not ideal, which makes it difficult for the onboard IMU to obtain the accurate yaw angle, and the motion capture system is difficult to be arranged in a non-laboratory environment. Therefore, the blimp designed in this paper focuses more on unidirectional maneuverability, which means that the blimp can only move along a set single direction.

\begin{figure}[h]
    \centerline{\includegraphics[width=0.4\textwidth]{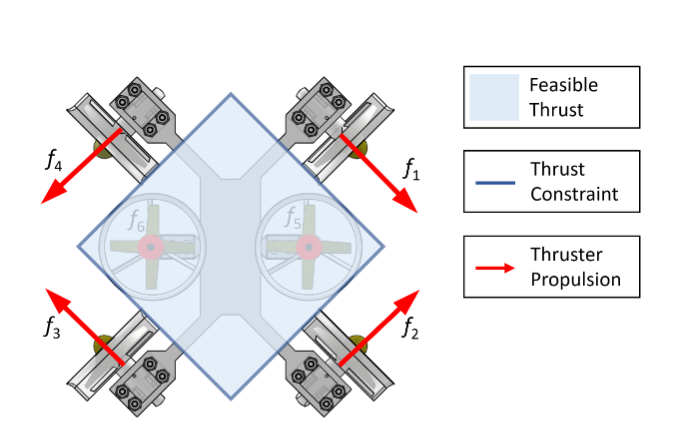}}
    \caption{Top view of the GT-MAB gondola. The horizontally mounted thrusters form an X-shaped conﬁguration to achieve symmetric holonomic actuation.}
    \label{fig:tview}
\end{figure}

The indoor miniature autonomous blimp has a unique thruster configuration that allows for a symmetrical drive, where each motor only needs to be propelled forward to achieve unidirectional drive. Meanwhile, omnidirectional drive can be realized after obtaining precise position and heading information.
The blimp pod contains four horizontally mounted propellers and two vertically mounted propellers (as shown in Figure \ref{fig:tview}), forming an X-shaped structure, so that the blimp can be symmetrically driven for planar motion. In addition, since the indoor blimp is a low-speed aerial robot, it does not need to change the direction of rotation frequently. This allows the motor to have higher efficiency, longer lifespan, and faster response time\cite{swing}.

\subsection{Coordinate Frame Systems}
\begin{figure}[htbp]
    \centerline{\includegraphics[width=0.3\textwidth]{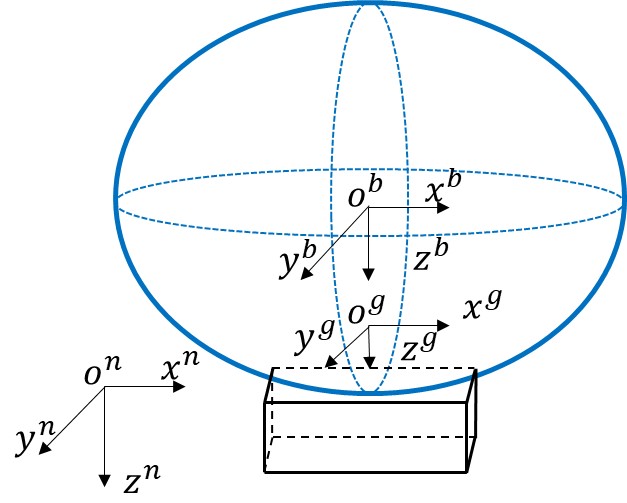}}
    \caption{Coordinate frame systems of the blimp.}
    \label{fig:frames}
\end{figure}

To facilitate understanding, we will firstly define the essential symbols and coordinate frame systems that are discussed in this chapter.
We consider $(\cdot)^n$ as the inertial frame. $(\cdot)^b$ is the body-fixed frame attached at the center of buoyancy, and $(\cdot)^g$ is another body-fixed frame attached at the center of gravity separately. The notation $(\cdot)_{b/n}$ represents the body-fixed frame kinematics variable of the vehicle attached at the center of buoyancy and expressed in the inertial frame. The direction of gravity $g$ aligns with the positive z-axis of the inertial frame. For a detailed depiction of the relationship between the body-fixed frame and the inertial frame, refer to Figure \ref{fig:frames}. In our notation, the pose of the vehicle fixed in the center of buoyancy in an inertial system as $\boldsymbol{\eta^{n}_{b/n}} = [ \boldsymbol{p^{b}_{b/n}} , \boldsymbol{\theta^{b}_{b/n}}]^\text{T}$. Here, $\boldsymbol{p^{b}} \in \mathbb{R}^3 $ represents the three-dimensional position, and $\boldsymbol{\theta^{b}_{b/n}} \in \mathbb{S}^3 $ represents the orientation angles of roll $\phi $, pitch $\theta$ , and yaw $\psi$. $\boldsymbol{v}^b_{b/n} \in \mathbb{R}^3$ and $\boldsymbol{\omega }^b_{b/n} \in \mathbb{R}^3$ are the linear and angular velocities of the blimp respectively.

\section{Underactuated Motion Model of Blimp}\label{2}
\begin{figure}[b]
    \centerline{\includegraphics[width=0.3\textwidth]{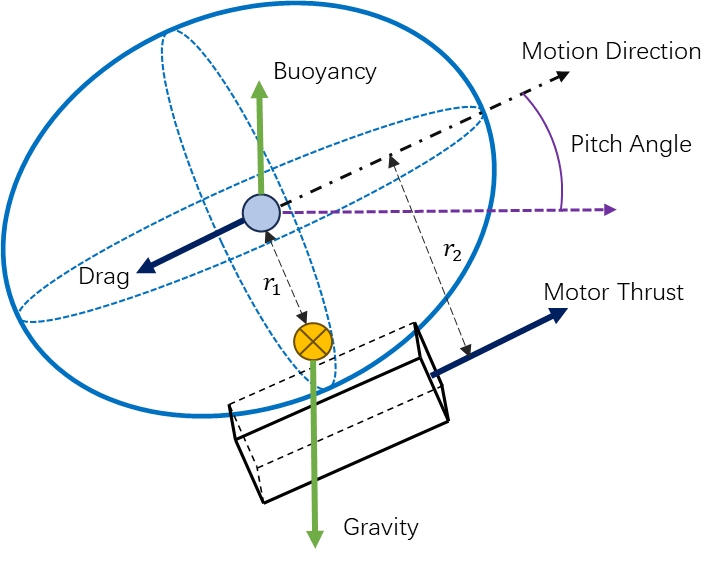}}
    \caption{Force analysis of Blimp in horizontal motion.}
    \label{fig:op}
\end{figure}

Building on our previous work \cite{identyfy}, which demonstrated the blimp's under-actuated structure and bottom-heavy design can generate an undesirable torque due to an offset between the propulsion and the air drag. This torque causes the blimp to pitch up during the horizontal motion. The forces on blimp during transitional motion in the horizontal plane is shown in Figure \ref{fig:op}. 
\begin{figure}[b]
    \centerline{\includegraphics[width=0.25\textwidth]{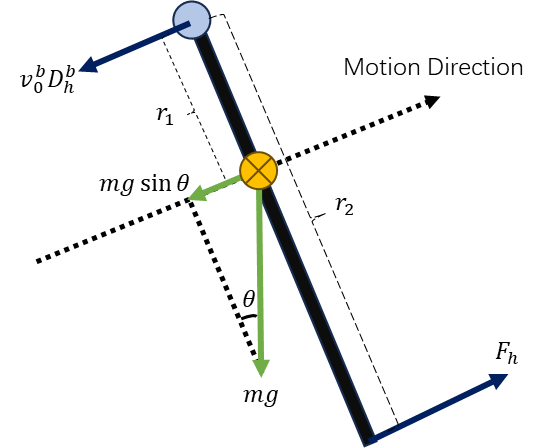}}
    \caption{The balanced forces of Blimp for compensated pitch-up torque.}
    \label{fig:res}
\end{figure}
In this Figure, $r_1$ denote the distances between the center of gravity and the center of buoyancy, and $r_2$ is the vertical distance between the thrusters and the center of buoyancy. 

Although such a pitch-up torque results in a nonlinear coupling between transitional and rotational motions. Our experiments\cite{swing} have revealed that when a blimp is moving horizontally, the combined effects of air drag $\boldsymbol{v}^b_{0}D^b_h$, the component gravity in the direction of motion $mg\sin{\theta_0}$ and motor thrust exerted on the gondola $F_h$ can reach a balance described as:
\begin{equation}
    F_h = \frac{r_1}{r_2} mg\sin{\theta_0} \label{eq:h_state}
\end{equation}
This balance only occurs when the blimp reaches a specific pitch angle $\theta_0$. And the effect has been visually represented in Figure \ref{fig:res}, illustrating the balanced forces that enable the blimp to maintain a stable transitional motion. 

$D^b_h$ represents the horizontal air drag coefficients of the Blimp, while $m$ represents the overall rigid-body mass. As a result, we can simplify the complete 6-DOF motion model to a 3-DOF motion model.

For 3-DOF vertical motion, it doesn't generate any rotational motion and horizontal motion. So, when the heave motion is the dominant movement, the Blimp is primarily subject to its own gravity $mg$, buoyancy, vertical thruster force $f^g_z$ and air drag $\boldsymbol{v}^z_{h,b/n} D^b_z$ in the vertical direction. Under this assumption, we can derive a model for the Blimp's vertical motion as follows:
\begin{equation}
    F^g_z = mg\boldsymbol{\dot{v}}^z_{h,b/n} + \boldsymbol{v}^z_{h,b/n} D^b_z \label{eq:v_state}
\end{equation}

In the context of 3-DOF horizontal motion, as illustrated in Figure \ref{fig:res}, the forces acting along the direction of motion are effectively cancelled out. This cancellation results in a decomposition of the remaining forces along the line connecting the center of buoyancy and the center of mass. These forces can be further separated into vertical and horizontal components. Considering this decomposition, we can derive the equation for the velocity in the horizontal direction, as denoted in equation:
\begin{equation}
    \dot{v}^b_{h,b/n,0} = \cos{\theta} (\dot{v}^b_0 + \frac{F_h}{m}) \label{eq:steady_state}
\end{equation}
And the equation for the velocity in the vertical direction as:
\begin{equation}
    \dot{v}^b_{z,b/n,0} = \sin{\theta} [\dot{v}^b_0  (1 - \frac{D^b_h}{m}) + \frac{F_h}{m}] \label{eq:steady_z_state}
\end{equation}
At the same time, when horizontal motion dominates, the propeller thrust, generated by a high-velocity airflow, introduces a hydro-static pressure difference between the top and bottom of the Blimp due to the Bernoulli effect. This pressure difference creates a nonlinear coupled motion of the blimp in both the horizontal and vertical directions. In other words, when the blimp is moving horizontally, there will always exists an additional force along the direction of gravity forcing the blimp to descend. Denoting this relationship as $\mathcal{B}(\cdot)$, we can derive the following equation:
\begin{equation}
    f^g_z = mg\boldsymbol{\dot{v}}^b_{z,b/n} + \mathcal{B}(\boldsymbol{v}^b_{h,b/n}) + \boldsymbol{v}^b_{z,b/n} D^b_z
    \label{eq:vb_state}
\end{equation}
By combining equations \eqref{eq:h_state}, \eqref{eq:v_state} ,\eqref{eq:steady_state}, \eqref{eq:steady_z_state} and \eqref{eq:vb_state}, we can derive the coupled a 3-DOF motion model of Blimp, represented by equation \eqref{eq:linear_motion}, in which $\textbf{p}^{b}_{b/n} = [p^{b}_{x,b/n}, p^{b}_{y,b/n}, p^{b}_{z,b/n}]^T$, $\textbf{f}^{g} = [f^{g}_{x}, f^{g}_{y}, f^{g}_{z}]^T$ and: 
\begin{align*}
        \textbf{B} = \begin{pmatrix} 1 & 0 & 0 \\ 0 & 1 & 0\\ \mathcal{B}(\cdot) & \mathcal{B}(\cdot) & 1 \end{pmatrix}
        \textbf{D} &= \begin{pmatrix} \frac{1}{m} & 0 & 0 \\ 0 & \frac{1}{m} & 0\\ 0 & 0 & -\frac{D^b_z}{m} \end{pmatrix} \\
        \textbf{M} = \begin{pmatrix} \frac{1}{m} & 0 & 0 \\ 0 & \frac{1}{m} & 0\\ 0 & 0 & \frac{1}{m} \end{pmatrix} 
        \textbf{0} &= \begin{pmatrix} 0 & 0 & 0 \\ 0 & 0 & 0\\ 0 & 0 & 0 \end{pmatrix} \\
\end{align*}
\begin{equation}
        \begin{pmatrix}    \dot{\textbf{p}}^{b}_{b/n}   \\    \ddot{\textbf{p}}^{b}_{b/n}   \end{pmatrix} = 
        \begin{pmatrix}    
                        \textbf{0} & \textbf{B} \\
                        \textbf{0} & \textbf{D} \\
        \end{pmatrix} 
        \begin{pmatrix}    \textbf{p}^{b}_{b/n}  \\    \dot{\textbf{p}}^{b}_{b/n} 
        \end{pmatrix} + 
        \begin{pmatrix}
            \textbf{0} \\
            \textbf{M}
        \end{pmatrix}
            \textbf{f}^g
        \label{eq:linear_motion}
\end{equation}



When the blimp experiences no horizontal or vertical displacement, it is considered to be in pure rotational mode. In this mode, $(\cdot)^g$ always moves in the same direction as the thrust force. Additionally, since the positions of $(\cdot)^b$ and $(\cdot)^g$ are parallel, $\omega^g = \omega^b $. Therefore, the motion model for pure rotation under the $(\cdot)^g$ can be described as follows:
\begin{equation}
    \tau_{z}^{g} = I_{z}^{g}\dot{\omega }_{z,g/n}^{b} +({D_{h}^{b}}{r_3}^2 + {D_{\omega {z}}^{b}}){\omega _{z,g/n}^{b}}
    \label{eq:rotation_motion}
\end{equation}
$I_{z}^{g}$ denotes the total rotational inertia at $(\cdot)^g$, $r_3$ is the distance between horizontally mounted thruster and the center of gondola, $\tau_{z}^{g}$ is the actuation around the z-axis, and $D_{\omega {z}}^{b}$ is the rotational air drag coefficient. $\dot{\phi} = \omega _z $, the pure motion model can be approximated in the inertial frame under:
\begin{equation}
    \begin{pmatrix}
        \dot{\phi} \\
        \ddot{\phi}
    \end{pmatrix} = 
    \begin{pmatrix}
        0 & 1 \\
        0 & -\frac{{D_{h}^{b}}{r_3}^2 + {D_{\omega {z}}^{b}}}{I_{z}^{g}}
    \end{pmatrix}
    \begin{pmatrix}
        \phi \\
        \omega_z
    \end{pmatrix} +
    \begin{pmatrix}
        0 \\
        \frac{1}{I_{z}^{g}}
    \end{pmatrix}
    \tau_{z}^{g}
\end{equation}

\section{Hybrid Controller Design}
The motion of a blimp is characterized by nonlinearities and a strong coupling between transitional and rotational movements. Our proposed balancing controller leverages the coupling between the horizontal velocity and pitch angle of the blimp, along with a stabilizing controller to swiftly mitigate the influences of external factors or human actions. We then introduce a preemptive mechanism that allows for a seamless transition between balanced and stabilized control during specific human interaction windows.






\subsection{Balancing Controller}
The balanced state of the blimp is shown in the Figure \ref{fig:res}. In this state, the blimp moves only in transition and does not rotate. Once the Blimp reaches this equilibrium state, any given horizontal velocity $v_{h,0}^b$ corresponds to a specific pitch angle $\theta_0$ that stabilizes it in this balanced state. Assuming that the blimp is flying stably with $v_{h,0}^b$ and $\theta_0$ under the PID velocity tracking controller. By linearizing the dynamics of the blimp in equation \eqref{eq:h_state} around the operating point, we can transform the motion model for velocity tracking into a motion model for angle tracking:
\begin{equation}
    m \Delta \dot{v}_h^b = D_h^b \Delta v_{h,b/n}^b  = \frac{r_1mg}{r_2}  \sin(\Delta\theta)
    \label{eq:angle_tracking}
\end{equation}
Considering that the coefficients in equation \eqref{eq:angle_tracking} are constant and that the relationship between $\sin \theta $ and $\theta $ can be approximated as linear when $\theta $ is not significantly large, we can simplify the control approach.
\begin{equation}
    \lim_{\theta \to 0}  \sin \theta = \theta + o(\theta) \approx \theta \label{eq:mac}
\end{equation}
This means that we do not need to measure any specific parameters of the blimp nor tracking the velocity of blimp by expensive motion capture equipment. Instead, we can focus on the the pitch angle when the blimp is moving at a constant speed. 
Based on our objectives, we proceed to design a state feedback controller that will effectively maintain the blimp's pitch angle at the desired operating point:
\begin{equation}
    \Delta f_h^g = -K_p\Delta \theta - K_i \int \Delta \theta dt - K_d \Delta \dot{\theta}
\end{equation}
Where $K_p$, $K_i$ and $K_d$ are the proportional, integral, and derivative gains, respectively. And $\Delta \theta $ and $\Delta \dot{\theta}$ represent the IMU's angular velocity and angular acceleration at the current reading, respectively. 

This controller will translate the specified planar direction inputs from the human pilot into corresponding pitch angles, and allowing the blimp to maintain the desired angle for execute the human's control commands.

\subsection{Stabilizing Controller}
To ensure the stability of the blimp while minimizing interference with the user's subsequent interactions, the stabilizing controller is designed to quickly bring the blimp to a state where $\omega_z = 0$.
To eliminate the pure rotational motion and bring the angular velocity to zero, we need to apply the control input as:
\begin{equation}
    \tau_{z}^{g} = \left\{\begin{matrix} 
        -u \text{ If } \omega_z  > 0  \\  
        +u  \text{ If }\omega_z < 0  \\
        0   \text{  If } \omega_z = 0
      \end{matrix}\right. 
\end{equation}
This type of control action is called a bang-bang controller because it switches between two extreme values $\pm u$ of the control input. The total energy of the system can be defined as:
\begin{equation}
    E = \frac{1}{2} I_{z}^{g} (\omega_{z,g/n}^{b})^2
\end{equation}
The rate of change of energy can be calculated as follows:
\begin{equation}
    \frac{dE}{dt} = I_{z}^{g} \omega_{z,g/n}^{b} \dot{\omega }_{z,g/n}^{b}
\end{equation}
Substituting the dynamics equation, we get:
\begin{equation}
    \frac{dE}{dt} = u \cdot (\omega_{z,g/n}^{b})^2
\end{equation}
Since our control input can take only two values ($+u$ or $-u$), in both cases, the worst-case scenario results in the maximum rate of change of energy. This implies that the bang-bang controller, which switches between $+u$ and $-u$ based on the sign of $\omega_{z,g/n}^{b}$, maximizes the rate of change of energy and hence minimizes the time required to bring the angular velocity to zero.

Therefore, the bang-bang controller is the optimal controller for eliminating the pure rotational motion of the blimp around the axis of gravity in the shortest possible time.

\subsection{Preemptive Mechanism}
Our proposed preemption mechanism ensures that the operator to feel that any new command they issue can immediately preempt the previous command being executed , while also allowing the blimp only switches user commands when it is in a balanced state.
\begin{figure}[h]
    \centering
    \includegraphics[width=0.45\textwidth]{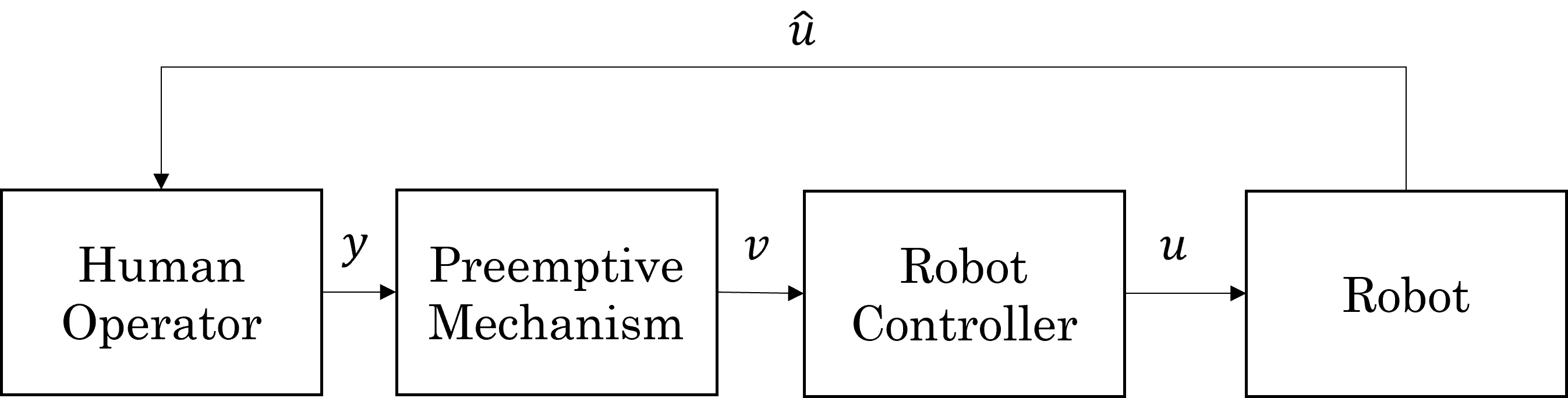}
    \caption{Block diagram of hybrid controller with preemptive mechanism.}
    \label{fig:Block}
\end{figure}
Based on previous research \cite{soon2008unconscious,libet1985unconscious,libet1993time,matsuhashi2008timing}, it has been established that the time interval between becoming aware of an intention is usually around 200 milliseconds. This means that the time from when the operator issues command $y_{t_i}$ to when the next command $y_{t_{i+1}}$ is issued based on the blimp's feedback $\hat{u}_{t_i}$ is no less than 200 milliseconds. Considering this, the preemption mechanism in Figure \ref{fig:Block} ensures that the duration of any $v_{t_i}$ is within 200 milliseconds, to ensure that the blimp is in an idle state when $v_{t_{i+1}}$ arrives. This approach ensures that the hybrid controller follows the consistency principle and enhances the operator's sense of control.

\begin{figure}[h]
    \centering
    \includegraphics[width=0.45\textwidth]{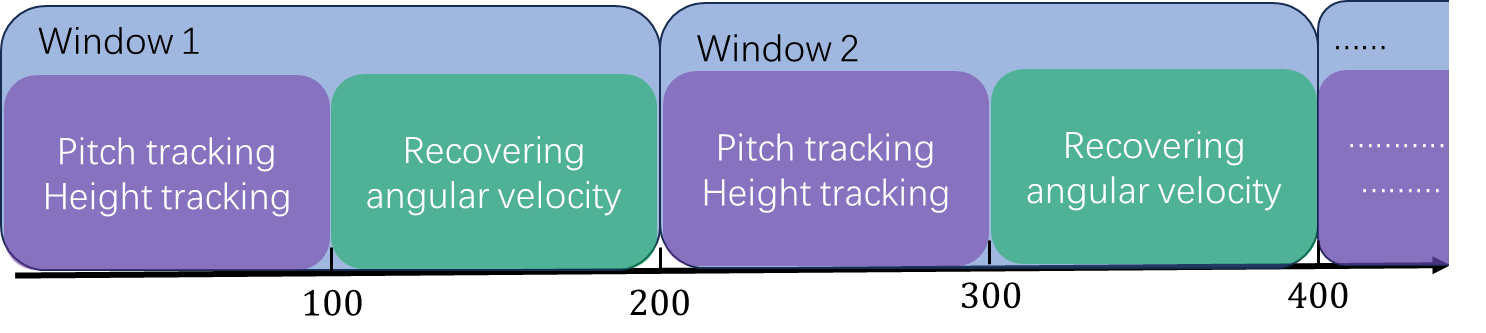}
    \caption{Human-computer interaction windows.}
    \label{fig:window}
\end{figure}
Additionally, as discussed in Chapter \ref{2}, the decoupling of the blimp's rotational and transitional kinematics depends on whether the blimp is in a balanced state. Therefore, to avoid swing on the $z^b$ axis during horizontal movement, the blimp should execute user commands when it is in a balanced state. 
Utilizing this property, we further refined the preemption mechanism, as shown in the Figure \ref{fig:window}. We divided the commands $v_{t_i}$ executed during the interaction time window into two parts. The first 100 milliseconds activate the balancing controller to maintain the blimp's movement to execute human commands. The second 100 milliseconds activate the stabilizing controller to make $\omega_z = 0$. Subsequently, the control is handed back to the human operator before the subsequent reaction time window commences.

\section{Experimental Results}
To evaluate the effectiveness of our hybrid controller in promoting human-assisted piloting, we conducted experiments in a controlled indoor testing environment. The participants' goal was to maneuver a blimp with human-assistive piloting functionality and a blimp without such functionality, using a joystick, from a fixed starting point through a target ring. During this process, we used an Opti-track system to record the blimp's angular velocity to observe the impact of our hybrid controller on the blimp's flight status.

\begin{figure}
	\centering
	\begin{subfigure}{1.0\linewidth}
		\includegraphics[width=\linewidth]{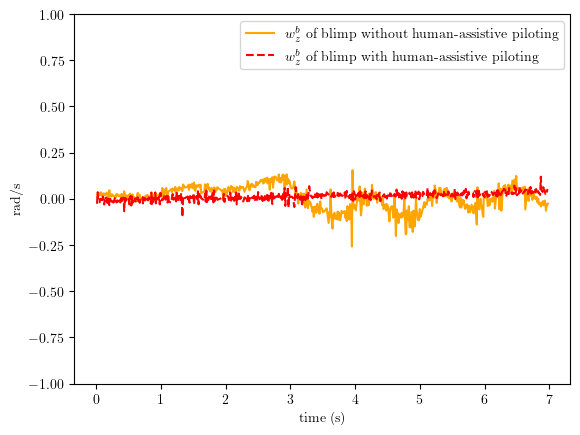}
		\caption{Participant 1}
	\end{subfigure}
    \vfill
	\begin{subfigure}{1.0\linewidth}
		\includegraphics[width=\linewidth]{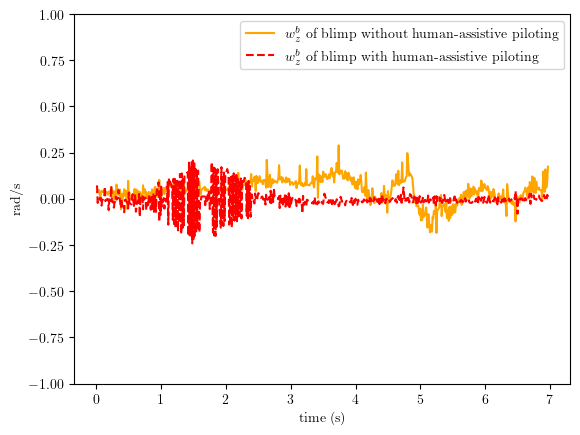}
		\caption{Participant 2}
	\end{subfigure}
     \vfill
	\begin{subfigure}{1.0\linewidth}
	        \includegraphics[width=\linewidth]{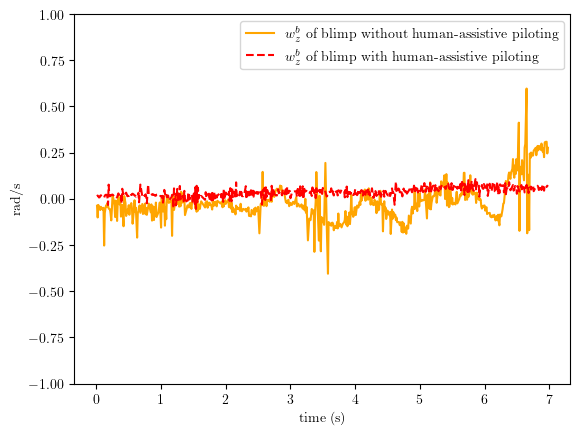}
	        \caption{Participant 3}
         \end{subfigure}
	\caption{$\omega_z^b$ of blimps during crossing experiments.}
	\label{fig:exz}
\end{figure}
As shown in the Figure \ref{fig:exz}, during the crossing experiment, the blimp equipped with the hybrid controller was able to dynamically maintain an angular velocity $\omega_z^b$ of 0 $\text{rad}/\text{s}$, effectively eliminating swing oscillation. On the other hand, the blimp without the hybrid controller experienced significant swing oscillation during the experiment which gradually accumulated.

\begin{figure}
	\centering
	\begin{subfigure}{1.0\linewidth}
		\includegraphics[width=\linewidth]{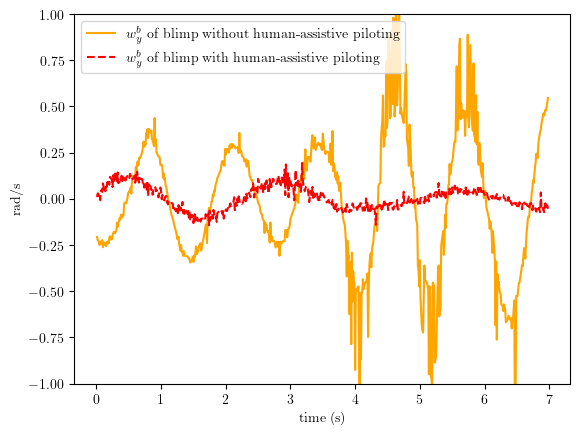}
		\caption{Participant 1}
	\end{subfigure}
    \vfill
	\begin{subfigure}{1.0\linewidth}
		\includegraphics[width=\linewidth]{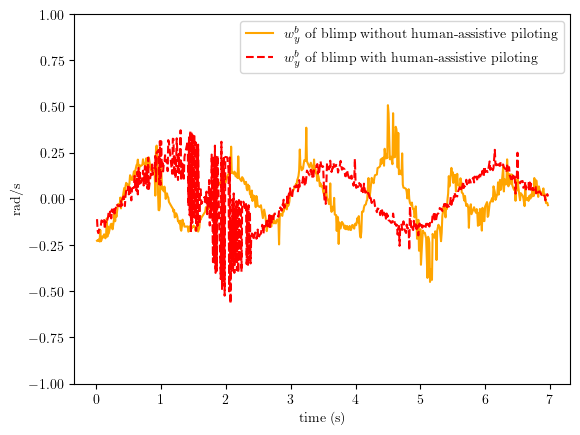}
		\caption{Participant 2}
	\end{subfigure}
     \vfill
	\begin{subfigure}{1.0\linewidth}
	        \includegraphics[width=\linewidth]{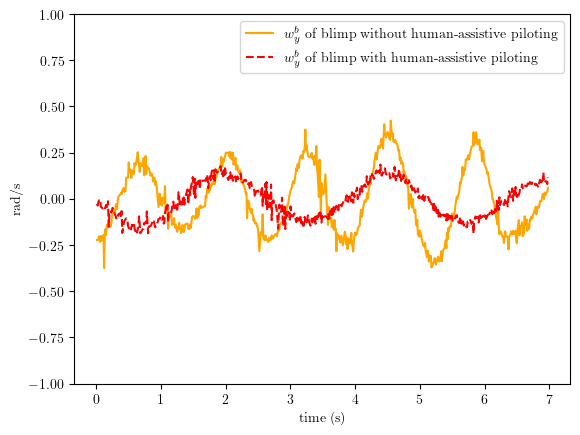}
	        \caption{Participant 3}
         \end{subfigure}
	\caption{$\omega_y^b$ of blimps during crossing experiments.}
	\label{fig:exy}
\end{figure}

Meanwhile, our experiments also analyzed the impact of the hybrid controller on the blimp's sway along the direction of travel. The experimental results shown in the Figure \ref{fig:exy}  indicate that with the assistance of the hybrid controller, the blimp operated by humans not only eliminated swing oscillation but also reduced the frequency and amplitude of sway along the direction of movement.

\section{Conclusion}
This paper introduces an innovative human-computer interaction technique for indoor miniature blimps. Due to the blimp's unique aerodynamic shape and underdriven design, its transitional and rotational motions are nonlinear and coupled. To address this challenge, we have developed closed-loop controllers with fast angular velocity corrections. These controllers effectively separate the coupled motions and account for the non-linearity.

By seamlessly switching between these two controllers within the human response window, our approach ensures that the blimp automatically stabilizes its attitude during the execution of the operator's commands without causing any disturbances. Through experimental validation, we demonstrate that our human-interactive flight controller enables precise maneuvering of the blimp indoors, reducing oscillations effectively.

Importantly, the approach presented in this paper is applicable to indoor micro blimps of various sizes. In future research, we plan to delve deeper into Human-Assistive Piloting and enhance the blimp's design by incorporating new on-board sensors.

\bibliographystyle{IEEEtran}
\bibliography{my} 

\end{document}